\documentclass[10pt,twocolumn,letterpaper]{article}

\usepackage{iccv}
\usepackage{times}
\usepackage{epsfig}
\usepackage{graphicx}
\usepackage{amsmath}
\usepackage{amssymb}
\usepackage{courier}
\usepackage{booktabs}
\usepackage{color,xcolor}
\usepackage{subfig}
\usepackage{multirow}
\usepackage{comment}
\usepackage{framed}
\usepackage{enumitem}
\usepackage{placeins} 
\usepackage{flushend}

\definecolor{lightgrey}{rgb}{0.97, 0.97, 0.97}
\definecolor{darkgreen}{RGB}{0, 140, 0}
\definecolor{antiquefuchsia}{rgb}{0.57, 0.36, 0.51}
\definecolor{auburn}{rgb}{0.43, 0.21, 0.1}
\newcommand{\rv}[1]{{\color{antiquefuchsia}[\textbf{Rv}:#1]}}

\setlist[itemize]{%
labelsep=5pt,%
labelindent=0.4\parindent,%
itemindent=0pt,%
leftmargin=*,%
itemsep=4pt,
topsep=0pt,
parsep=0pt,
partopsep=0pt
}

\setlist[enumerate]{%
labelsep=5pt,%
labelindent=0.4\parindent,%
itemindent=0pt,%
leftmargin=*,%
itemsep=3pt,
topsep=4pt,
parsep=0pt,
partopsep=0pt
}

\makeatletter
\renewcommand{\paragraph}{%
  \@startsection{paragraph}{4}%
  {\z@}{0.6em}{-1em}%
  {\normalfont\normalsize\bfseries}%
}

\usepackage[pagebackref=true,breaklinks=true,letterpaper=true,colorlinks,bookmarks=false]{hyperref}

\iccvfinalcopy 

\def\iccvPaperID{5006} 
\def\httilde{\mbox{\tt\raisebox{-.5ex}{\symbol{126}}}}

\ificcvfinal\fi
\begin{document}

\title{Billion-scale semi-supervised learning for image classification \vspace{-5pt}}



\author{I. Zeki Yalniz\\
\and
Herv\'e J\'egou \\
\and
Kan Chen \\
Facebook AI\\
\and
Manohar Paluri \\
\and
Dhruv Mahajan \\
}

\maketitle

\begin{abstract}
This paper presents a study of semi-supervised learning with large convolutional networks. We propose a  pipeline, based on a teacher/student paradigm, that leverages a large collection of unlabelled images (up to 1 billion). Our main goal is to improve the performance for a given target architecture, like ResNet-50 or ResNext.
We provide an extensive analysis of the success factors of our approach, which
leads us to formulate some recommendations to produce high-accuracy models for image classification with semi-supervised learning.
As a result, our approach brings important gains to standard architectures for image, video and fine-grained classification. For instance, by leveraging one billion unlabelled images, our learned vanilla ResNet-50 achieves 81.2\% top-1 accuracy on Imagenet benchmark.

\end{abstract}


\vspace{-10pt}
\section{Introduction}
\label{sec:introduction}

Recently, image and video classification techniques leveraging web-scale weakly supervised datasets have achieved state-of-the-art performance on variety of problems, including image classification, fine-grain recognition, etc. However, weak supervision in the form of tags has a number of drawbacks. There is a significant amount of inherent noise in the labels due to non-visual, missing and irrelevant tags which can significantly hamper the learning of models. Moreover, weakly supervised web-scale datasets typically follow a Zipfian or long-tail label distribution 
that favors good performance only for the most prominent labels. Lastly, these approaches assumes the availability of large weakly supervised datasets for the target task, which is not the case in many applications.

\begin{figure}
\vspace{-5pt}
\includegraphics[width=\linewidth]{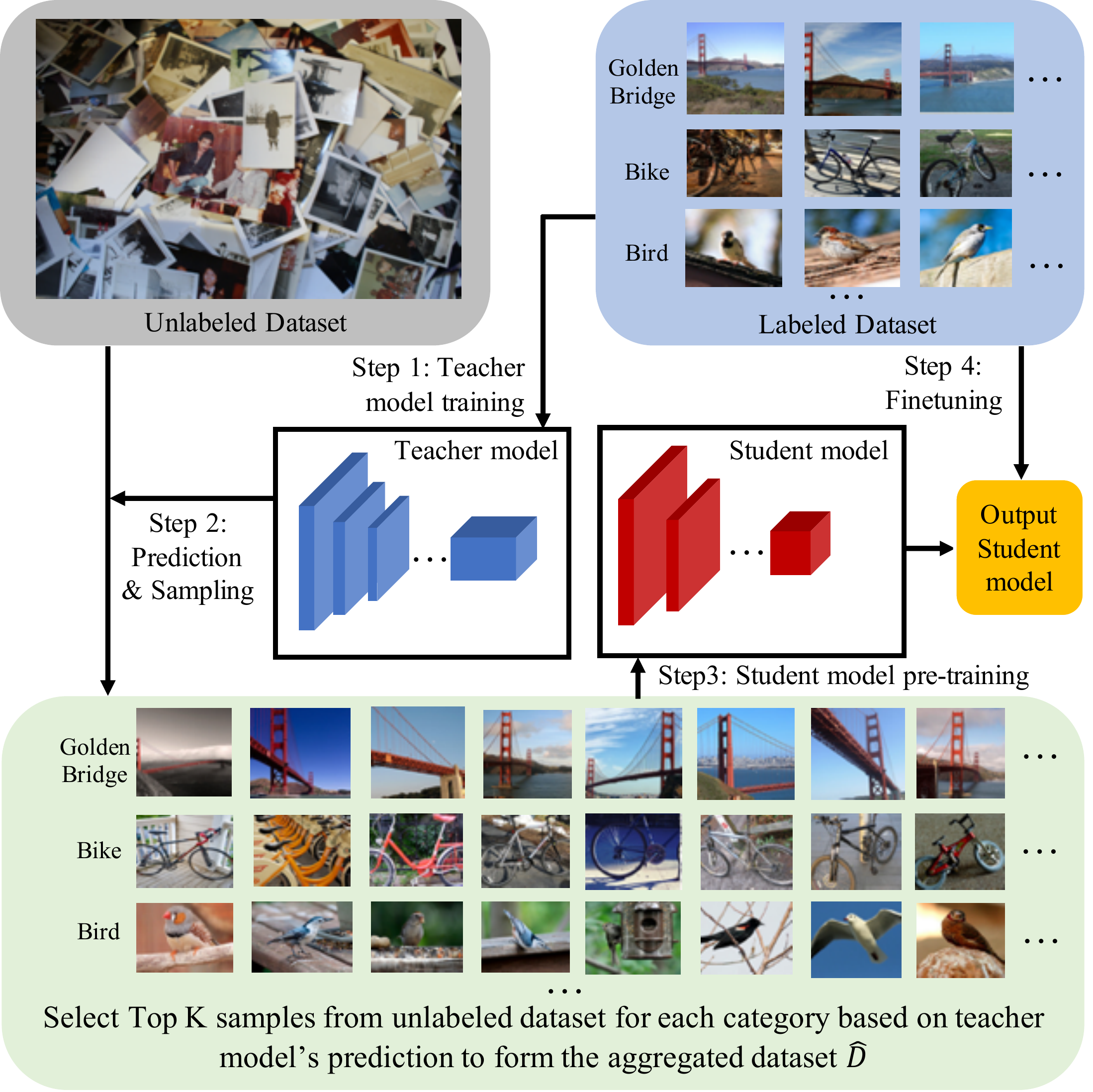}
\vspace{-20pt}
\caption{Illustration of our approach: with a strong teacher model, we extract from a very large unlabelled image collection (100M--1 billion images) a new (large) training set. The student model is first trained with this noisy  supervision, and fine-tuned with the original dataset.
\label{fig:splash}}
\vspace{-15pt}
\end{figure}



This paper explores web-scale semi-supervised deep learning. We leverage billions of unlabeled images along with a relatively smaller set of task-specific labeled data. To the best of our knowledge, semi-supervised learning with neural networks has not been explored before at this scale. In order to scale our approach, we propose the simple semi-supervised strategy depicted in Figure~\ref{fig:splash}: (1) We train on the labeled data to get an initial teacher model; (2) For each class/label, we use the 
predictions of this teacher model to rank the unlabeled images and pick top-K images to construct a new training data; (3) We use this data to train a student model, which typically differs from the teacher model: hence we can target to reduce the complexity at test time; (4) finally, pre-trained student model is fine-tuned on the initial labeled data to circumvent potential labeling errors.

Compared to most weakly-supervised approaches, we circumvent the issue of long-tail distribution by selecting same number of images per label. This is possible since for a tail class like ``African Dwarf Kingfisher'' bird, there may not be sufficient number of weakly-supervised/tagged examples, but a lot of un-tagged images of this bird is likely to exist in the unlabelled dataset. A teacher model trained with labels is able to identify enough images from the unlabeled data and hence to improve the recall for tail classes.

\begin{table*}[t]
\vspace{-8pt}
\fcolorbox{black}{lightgrey}{\small
\begin{minipage}{0.98\linewidth}
\centering
\caption{OUR RECOMMENDATIONS FOR LARGE-SCALE SEMI-SUPERVISED LEARNING\label{tab:recommendations}}
\label{tab:recipe}
\vspace{-8pt}
\begin{tabular}{lp{0.4\linewidth}}
\textbf{1.} &
Train with a teacher/student paradigm:
It produces a better model for a fixed complexity, even if the student and teacher have the same architecture. \\
\textbf{2.} &
Fine-tune the model with true labels only. \\
\textbf{3.} &
Large-scale unlabelled dataset is key to performance. \\
\end{tabular}
\hfill
\begin{tabular}{lp{0.4\linewidth}}
\textbf{4.} &
Use a large number of pre-training iterations (as opposed to a vanilla supervised setting, where a number of epochs as used in common practice is sufficient).  \\
\textbf{5.} &
Build a balanced distribution for inferred labels. \\
\textbf{6.} &
Pre-training the (high-capacity) teacher model by weak supervision (tags) further improves the results.
\end{tabular}
\end{minipage}
}
\vspace{-10pt}
\end{table*}

The choice of the semi-supervised algorithm described above is motivated by several approaches, such as self-training, distillation, or boosting. In fact, under some specific settings of our teacher and student models, and of the type of data, our work boils down to these approaches.
Nevertheless, our paper analyzes several variants, such as considering a pure self-training learning procedure by removing the teacher/student model, or fine-tuning on the inferred labels instead of only those from the labelled set.


Our reference pipeline trains a teacher model that is more powerful than the student model. 
Even in the case of self-training --when teacher and student models are the same--, the sheer scale of unlabeled data allows us to achieve significant gains. Our analysis shows the performance is sensitive to several factors like strength of initial (teacher) model for ranking, scale and nature of unlabeled data, relationship between teacher and final model, etc. 
In particular, we analyze the performance as a function of relative strength of teacher and student models.
We also show that leveraging hashtags or queries in search as a weak-supervision signal to collect the unlabeled dataset significantly boosts the performance.


Overall, this paper makes the following contributions:
\begin{itemize}
\item We explore the semi-supervised deep-learning at a large scale of billions of unlabeled examples and show that a simple strategy of ranking unlabeled images with a teacher model trained on labeled data works is effective for learning a powerful (student) network. 

\item We analyze the conditions under which such a strategy gives a substantial benefit.
Inspired by a prior work~\cite{perronnin2012towards} on large-scale supervised learning, we summarize these recommendations in Table~\ref{tab:recommendations}.

\item More specifically, we present a detailed analysis and ablations of various design choices, like strength of student and teacher CNN models, nature of unlabeled dataset, number of examples selected per label, etc.
\item We demonstrate the performance of our method on popular classification benchmarks for both images and videos and significantly outperforms the state of the art.
\item Our proposal is also effective in other tasks, namely video classification and fine-grain recognition.
\end{itemize}

This paper is organized as follows. Section~\ref{sec:related} introduces related works and terminology.
Section~\ref{sec:approach} introduces our approach. 
Section~\ref{sec:exp_imagenet} provides an extensive analysis and ablation studies for the classification task. Other tasks are evaluated in Section~\ref{sec:exp_video_transfer}.
Section~\ref{sec:conclusion} concludes the paper.


\section{Related Work}
\label{sec:related}

In this section we review several topics and specific works related to our approach. Since the terminology is not always consistent across papers, we clarify the different concepts employed through our paper. 

\paragraph{Image classification.} Our work mainly targets at improving image classification, whose performance is routinely evaluated on benchmarks like Imagenet~\cite{deng2009imagenet}. In the recent years, a lot of effort has been devoted to improving the neural networks architectures. For instance the introduction of residual connections~\cite{he2016deep} has led to several competitive convolutional neural networks~\cite{hu2018squeeze,huang2017densely,xie2017aggregated} that are widely adopted by the community. 
More recently, other directions have been investigated to further push the state of the art~\cite{soa_imagenet}
One proposal is to learn a gigantic convolutional neural network taking high-resolution images on input~\cite{huang2018gpipe}, as resolution is a key factor to classification performance~\cite{He2016IdentityMI}. Another line of research~\cite{mahajan2018exploring} learns a high-capacity convnet with a large weakly labeled dataset, in which annotation is provided in the form of tags.

\paragraph{Transfer learning} is a task in which one re-uses a network trained on a large labelled corpus to solve other tasks for which less data is available. This is a historical line of research in machine learning, see the survey~\cite{pan2010survey} by Pan and Yang for an extensive classification of the different problems and approaches related to this topic. 
This approach is typically required for tasks for which no enough labels are available~\cite{oquab2014learning}, such a fine-grain classification or detection (although a recent work~\cite{he2018rethinking} shows that learning from scratch is possible in the latter case). 

On a side note, transfer learning is employed to evaluate the quality of \textbf{unsupervised learning} ~\cite{dosovitskiy2014discriminative}: a network is trained with unlabelled data on a proxy task, and the quality of the trunk is evaluated by fine-tuning the last layer on a target task. 
A particular case of transfer is the task of \textbf{low-shot learning}, where one has to learn with very few examples of a new class. Typical works on this line of research propose parameter prediction~\cite{BHVTV16,qi2018low} and data augmentation~\cite{gao2018low,bharath2017low}.

\paragraph{Semi-supervised learning.} Recent approaches have investigated the interest of using additional unlabelled data to improve the supervision. To the best of our knowledge, most results from the literature obtained mitigated results~\cite{mahajan2018exploring}, except those employing weak labels for the additional data, which is a form a weak supervision. 

Recent approaches~\cite{douze2018low,Rohrback2013transfer} revisited traditional label propagation algorithms (See \cite{chapterLabelProp06,DB13} for a review) to leverage the large YFCC unlabelled collection for low-shot learning. Douze \etal~\cite{douze2018low} improve classification when the number of labelled data for the new classes is very low (1--5), but the gain disappears when more labelled data is available. Additionally this strategy has an important limitation: it requires to compute a large graph connecting neighboring images, which is constructed with an adhoc descriptor. 
This class of approach is often referred to as \textbf{transductive methods}: a few labels are provided and the goal is to extend the labeling to the unlabeled data. 
A classifier must be learned to provide an out-of-sample extension~\cite{CWSS07,JSL15}. 


\paragraph{Distillation} was originally introduced to compress a large model~\cite{hinton2015distilling}, called the \emph{teacher}, into a smaller one. 
The distillation procedure amounts to training the small model (the \emph{student}) such that it reproduces at best the output of the teacher. The student model is typically smaller and faster than the teacher model. Distillation can be seen as a particular case of \textbf{self-training}, in that the teacher model makes prediction on unlabelled data, and the inferred labels are used to train the student in a supervised fashion. 
This strategy was shown successful for several tasks, like detection~\cite{radosavovic2018data}. On a related note, back-translation~\cite{sennrich2016improving} is key to performance in translation~\cite{lample2018phrase}, offering large gains by leveraging (unannotated) monolingual data. 

To the best of our knowledge, our paper is the first to demonstrate that and how it can be effective for image classification by leveraging a large amount of unannotated data. 


\paragraph{Data augmentation} is a key ingredient for performance in image classification, as demonstrated by Krizhevsky \etal~\cite{krizhevsky2012imagenet} in their seminal paper. Several recent strategies have been successful at improving classification accuracy by learning the transformation learned for augmentation~\cite{cubuk2018autoaugment,paulin2014transformation} or by synthesizing mixtures of images~\cite{zhang2017mixup}. 
To some extent, our semi-supervised approach is a special form of data-augmentation, in that we add new actual images from a supplemental unlabelled dataset to improve the performance. 



\section{Our semi-supervised training pipeline}
\label{sec:approach}
This section introduces our reference large-scale semi-supervised approach. 
This algorithm builds upon several concepts from the literature. We complement them with good practices and new insights gathered from our extensive experimental study to achieve a successful learning.
We then discuss several key considerations regarding our approach and provides some elements of intuition and motivation supporting our choices. 

\subsection{Problem statement: semi-supervised learning}
We consider a multi-class image classification task, where the goal is to assign to a given input image a label. We assume that we are given a labeled set $\mathcal D$ comprising $N$ examples with corresponding labels.
In particular we will consider the Imagenet ILSVRC classification benchmark, which provides a labeled training set of $N$\,=\,1.2 million images covering 1000 classes.

In addition, we consider a large collection $\mathcal U$ of $M \gg N$ unlabeled images. Note that the class distribution of examples in $\mathcal U$ may be different from those in $\mathcal D$.
Our semi-supervised learning task extends the original problem by allowing to train the models on $\mathcal D \cup \mathcal U$, \ie, to exploit unlabelled data for the learning. The inference, and therefore the performance evaluation, is unchanged, and follow the protocol of the target task. %

%

\subsection{Our learning pipeline}
\label{sec:pipeline}
Our semi-supervised approach consists of four stages: \smallskip

\noindent \fcolorbox{black}{lightgrey}{
\begin{minipage}{0.95\linewidth}
\begin{enumerate}[label=(\arabic*), ref=\arabic*]
\item Train a teacher model on labeled data $\mathcal D$;
\item Run the trained model on unlabeled data $\mathcal U$ and select relevant examples for each label to construct a new labeled dataset $\hat{\mathcal D}$;
\item Train a new student model on $\hat{\mathcal D}$;
\item Fine-tune the trained student on the labeled set $\mathcal D$. 
\end{enumerate}
\end{minipage}
}  \smallskip

This pipeline is akin to existing distillation works. We depart from the literature in (i) how we jointly exploit unlabelled and labelled data; (ii) how we construct $\hat{\mathcal D}$; (iii) the scale we consider and targeting improvement on Imagenet.
We now discuss these stages into more details.

\paragraph{Teacher model training.}

We train a teacher model on the labeled dataset $\mathcal D$ to label the examples in $\mathcal U$. The main advantage of this approach is that inference is highly parallelizable. Therefore we perform it on billions of examples in a short amount of time, either on CPUs or GPUs.
A good teacher model is required for this stage, as it is responsible to remove unreliable examples from $\mathcal U$ in order to label enough relevant examples correctly without introducing substantial labelling noise.

\begin{figure*}[t]
\vspace{-10pt}
\centering
\includegraphics[width=0.8\linewidth]{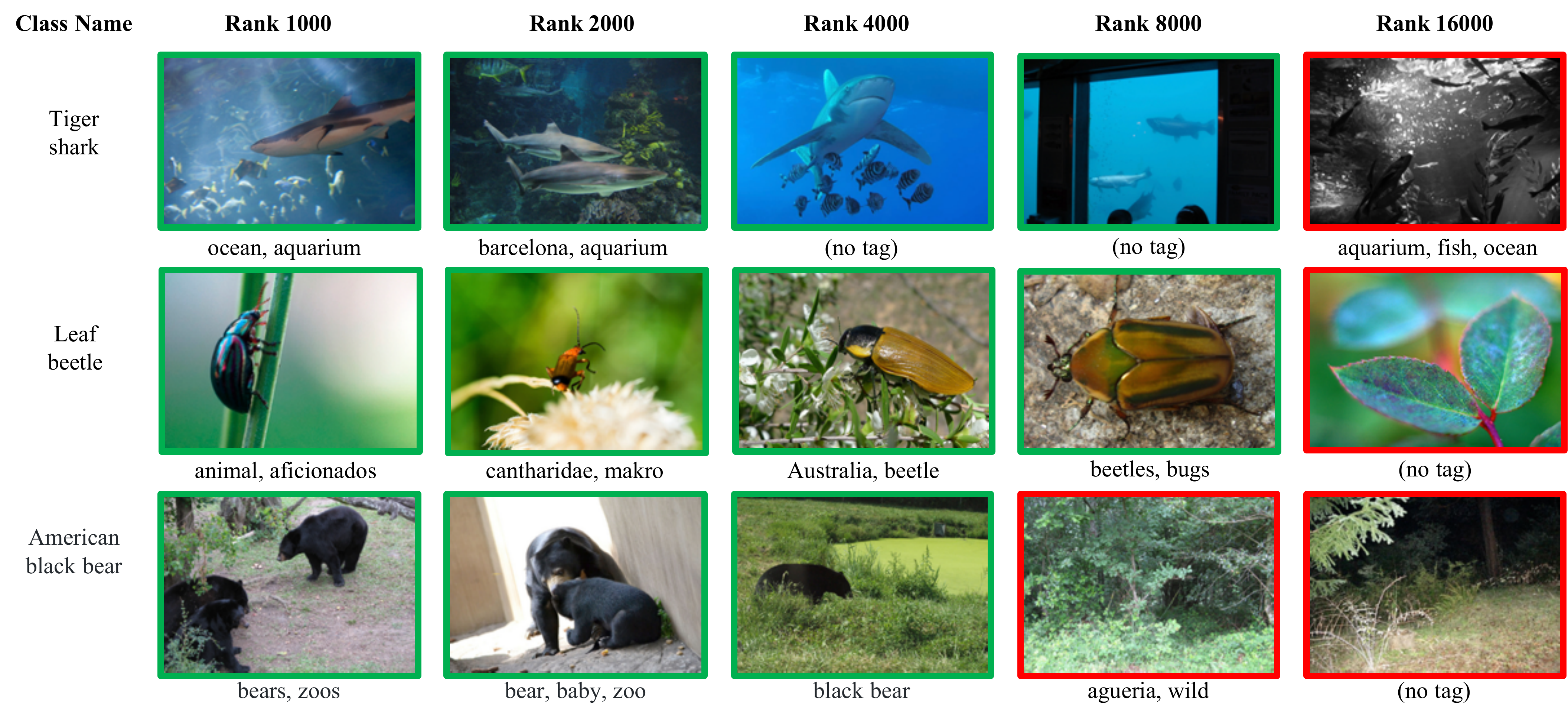}
\vspace{-10pt}
\caption{Examples of images from YFCC100M collected by our procedure for the classes ``Tiger shark'', ``Leaf beetle'' and ``American black bear'' for a few ranks.
\label{fig:examples_collected}}
\vspace{-10pt}
\end{figure*}

\paragraph{Data selection and labeling.}
Our algorithm leverage the desirable property that learning is tolerant to a certain degree of noise~\cite{natarajan2013noisylabels}. Our goal here is to collect a large number of images while limiting the labelling noise.
Due to large scale of unlabeled data, a large proportion of data may not even contain any of the classes in $\mathcal I$. We propose to select {\it{top-K}} examples from $\mathcal U$ for each target label. First, the teacher model is run on each example in $\mathcal U$ to obtain the softmax prediction vector. For each image, we retain only the classes associated with the $P$ highest scores, $P$ being a parameter accounting for the fact that we expect only a few number of relevant classes to occur in each image. The reason for choosing $P$\,$>$\,$1$ is that it is difficult to identify accurately under-represented concepts, or some may be occulted by more prominent co-occurring concepts. The role of this parameter is analyzed into more details in Section~\ref{sec:exp_imagenet}.

Then, for each class $l$, we rank the images based on the corresponding classification scores. The top--$K$ images define the set of $\hat{\mathcal D}_l$ of images that are henceforth considered as new positive training examples for that class. We also experiment with using raw predictions before applying the softmax for ranking, but found that they perform worse (see supplementary material).
The new image training set collected for  the multi-class classification problem is therefore defined as
\vspace{-5pt}
$$
\hat{\mathcal D} = \bigcup\limits_{l=1}^{L} \hat{\mathcal D}_l.
\vspace{-5pt}
$$

Figure~\ref{fig:examples_collected} shows the qualitative results of ranking YFCC100M dataset~\cite{thomee2016yfcc100m} with ResNet-50~\cite{he2016deep} teacher model trained on ImageNet-val~\cite{russakovsky2015imagenet} for $5$ classes. As expected, the images at the top of ranking are simple and clean without much labelling noise. They become progressively less obvious positives as we go down in the ranking. Introducing such hard examples is important to get a good classifier, but then the probability of introducing false positive becomes higher as $K$ increases.
Therefore, there is an important trade-off on $K$ that strongly depends on the ratio $K/M$. It is analyzed in Section~\ref{sec:exp_imagenet}. There is also a subtle interplay between $P$, $K$ and $M$. As mentioned above, setting $P>1$ is a way to collect enough reliable examples for the tail classes when the collection ${\mathcal U}$ is not large enough.


\paragraph{Student model training and fine-tuning.}
We train the student model using the supervision provided by $\hat{\mathcal D}$.
While our main interest is to learn a more simple model than the teacher, it is actually possible to use the same architecture for the teacher and the model, and as we will see later in the paper there is a practical interest.
Note that, although our assignment procedure makes it possible to assign an image to multiple classes in $\hat{\mathcal D}$, we still treat the problem as multi-class classification by simply allowing the image to be replicated in the dataset. The trained student model is then fine-tuned on $\mathcal D$ and evaluated on the test data.
Softmax loss is used for both pre-training and fine-tuning. 

\noindent \textit{Remark:} It is possible to use a mixture of data in $\mathcal D$ and $\hat{\mathcal D}$ for training like in previous approaches~\cite{radosavovic2018data}. However, this requires for searching for optimal mixing parameters, which depend on other parameters. This is resource-intensive in the case of our large-scale training.
Additionally, as shown later in our analysis, taking full advantage of large-scale unlabelled data requires adopting long pre-training schedules, which adds some complexity when mixing is involved.

Our fine-tuning strategy on $\mathcal D$ does not have these problems: instead it requires a parameter sweep over the initial learning rate for fine-tuning, which is more efficient due to smaller data size and training schedules. It also ensures that the final model is fine-tuned with clean labels.



\subsection{Bonus: weakly-supervised pre-training}

We consider a further special case where in addition to $\mathcal U$, $\mathcal D$ also includes weakly-supervised data ($\mathcal W$) plus a relatively small task-specific data ($\mathcal V$). Prior work~\cite{mahajan2018exploring} has shown that the teacher model obtained by pre-training on $\mathcal W$ followed by fine-tuning on $\mathcal V$ is a compelling choice. The student model obtained by training on the data selected by the teacher model is significantly better than the one obtained by training directly on the weakly-supervised data. As shown later in the experiments, this particular setting allows us to achieve state-of-the-art results.


\section{Image classification: experiments \& analysis}
\label{sec:exp_imagenet}

The effectiveness of our approach is demonstrated for the image classification task by performing a series of experiments on the ImageNet1K dataset. These experiments support the main recommendations proposed in Table~\ref{tab:recommendations}.

\subsection{Experimental Setup}
\label{sec:setup}

\paragraph{Datasets:} The following web-scale datasets are used for semi-supervised learning experiments involving an unlabeled dataset $\mathcal U$.
\begin{itemize}
\item {\it{YFCC-100M}}~\cite{thomee2015new} is a publicly available dataset of about $90$ million images from Flickr website with associated tags. After removing duplicates, we use this data for most experiments and ablations. 

\item {\it{IG-1B-Targeted: }}Following~\cite{mahajan2018exploring}, we collected a dataset of 1B public images with associated hashtags from a social media website. 
We consider images tagged with at least one of the $1500$ hashtags associated with one of the $1000$ ImageNet-1k classes.

\end{itemize}
Unless specified otherwise, we use the standard ImageNet with $1000$ classes as the labeled set $\mathcal D$.
\vspace*{0.05in}\\
\noindent {\bf{Models: }}For student and teacher models, we use residual networks~\cite{he2016deep}, ResNet-d with $d$\,=\,$\{18, 50\}$ and residual networks with group convolutions~\cite{xie2017aggregated}, ResNeXt-101 32X{\it{C}}d with $101$ layers and group widths $C$\,=\,$\{4, 8, 16, 48\}$. In addition, we consider a ResNeXt-50 32x4 network. The number of model parameters is shown in Table~\ref{tab:yfcc_teacher_capacity}.

\paragraph{Training Details:} We train our models using synchronous stochastic gradient descent (SGD) on $64$ GPUs across $8$ machines. Each GPU processes $24$ images at a time and apply batch normalization~\cite{ioffe2015batch} to all convolutional layers on each GPU. The weight decay parameter is set to $0.0001$ in all the experiments. We set the learning rate following the linear scaling procedure proposed in~\cite{goyal2017imagenethour} with a warm-up and overall minibatch size of $64 \times 24 = 1536$.

For pre-training, we use a warm-up from $0.1$ to $0.1/256 \times 1536$, where $0.1$ and $256$ are the standard learning rate and minibatch size used for ImageNet training. We decrease the learning rate by a factor of $2$ at equally spaced steps such that the total number of reductions is $13$ over the course of training. For fine-tuning on ImageNet, we set the learning rate to $0.00025/256 \times 1536$ and use the learning rate schedule involving three equally spaced reductions by a factor of 0.1 over $30$ epochs.

\paragraph{Default parameters.} In rest of this section, the number of "iteration" refers to the total number of times we process an image (forward/backward). Unless stated otherwise, we adopt the following parameters by default: We process \#iter\,=\,1B images for training. We limit to $P$\,=$\,10$ the number of concepts selected in an image to collect a short-list of $K$\,=\,16k images per class for ${\hat{\mathcal D}}$. Our default teacher model is a vanilla ResNext-101 architecture (32x48) as defined by Xie \etal~\cite{xie2017aggregated}. Since our main goal is to maximize the accuracy of our student model, we adopt a vanilla ResNet-50 student model~\cite{he2016deep} in most experiments to allow a direct comparison with many works from the literature. Finally, the unlabeled dataset ${\mathcal U}$ is YFCC100M by default.

\subsection{Analysis of the different steps}
\label{sec:stepanalysis}
%

\begin{table}[t]
\centering{\small
\begin{tabular}{@{}l|c|c|c@{}}
\toprule
\multirow{2}{*}{Student Model} &  \multicolumn{2}{|c|}{Ours: Semi-supervised} &  Fully\\ 
                     & Pre-training    &  Fine-tuned  &  Supervised \\
\midrule
ResNet-18            & 68.7            & \textbf{72.6}     & 70.6  \\
ResNet-50            & 75.9            & \textbf{79.1}     & 76.4 \\
\midrule
ResNext-50-32x4      & 76.7            & \textbf{79.9}     & 77.6 \\
ResNext-101-32x4     & 77.5            & \textbf{80.8}     & 78.5 \\
ResNext-101-32x8     & 78.1            & \textbf{81.2}     & 79.1 \\
ResNext-101-32x16    & 78.5            & \textbf{81.2}     & 79.6 \\
\bottomrule
\end{tabular}}
\vspace{-5pt}
\caption{ImageNet1k-val top-1 accuracy for students models of varying capacity 
before and after fine-tuning compared to corresponding fully-supervised baseline models.
\label{tab:yfcc_student_capacity}}
\end{table}

\paragraph{Our approach vs supervised learning.}
Table~\ref{tab:yfcc_student_capacity} compares our approach with the fully-supervised setting of training only on ImageNet for different model capacities. Our teacher model brings a significant improvement over the supervised baseline for various capacity target models (1.6--2.6\%), which supports our recommendation~in Table~\ref{tab:recipe}.

\paragraph{Importance of fine-tuning.} Since labels of pre-training dataset $\hat{\mathcal D}$ and labeled dataset $\hat{D}$ are same, we also compare the performance of model trained on $\hat{D}$ before and after the last full fine-tuning step of our approach. From  Table~\ref{tab:yfcc_student_capacity} we observe that fine-tuning the model on clean labeled data is crucial to achieve good performance (Point 2 in Table~\ref{tab:recipe}).

\begin{table}[t]
\centering {\small
\begin{tabular}{l@{}rc|c|c} 
\toprule
\multicolumn{3}{c|}{Teacher} & Student & \multirow{2}{*}{Gain (\%)} \\
Model  & \# Params  & top-1 & top-1  &  \\ 
\midrule
ResNet-18         &  8.6M  & 70.6  & 75.7	 & -0.7 \\ 
ResNet-50         &  25M   & 76.4  & 77.6  & +1.2 \\ 
ResNext-50-32x4   &  25M   & 77.6 & 78.2  & +1.8 \\ 
ResNext-101-32x4  &  43M   & 78.5  & 78.7  & +2.3 \\ 
ResNext-101-32x8  &  88M   & 79.1 & 78.7  & +2.3 \\ 
ResNext-101-32x16 &  193M  & 79.6 & 79.1  & +2.7 \\ 
ResNext-101-32x48 &  829M  & 79.8 & 79.1  & +2.7 \\ 
\bottomrule
\end{tabular}}
\vspace{-5pt}
\caption{Varying the teacher capacity for training a ResNet-50 student model with our approach. The gain is the absolute accuracy improvement over the supervised baseline.
}
\label{tab:yfcc_teacher_capacity}
\end{table}

\paragraph{Effect of the student and teacher capacities.}
In Table~\ref{tab:yfcc_student_capacity}, we observe that the accuracy increases significantly for lower capacity student models, before saturating around the ResNeXt-101 32x8 model.
Table~\ref{tab:yfcc_teacher_capacity} shows accuracy as a function of teacher model capacity. The accuracy of ResNet-50 student model improves as we increase the strength of teacher model until ResNeXt-101 32x16. Beyond that the classification accuracy of the teacher saturates due to the relatively small size of ImageNet. Consequently, increasing the teacher model capacity further has no effect on the performance of the student model.


Together, the above experiments suggest that the improvement with our approach depends on the relative performance of student and teacher models on the original supervised task (\ie, training only on the labeled data $\mathcal D$). Interestingly, even in a case of where both teacher and student models are ResNet-50, we get an improvement of around $1\%$ over the supervised baseline.

\begin{table}[t]
\centering {\small
\begin{tabular}{c|cc|c|ccc}
\toprule
      & \multicolumn{2}{c|}{ResNet-*} & ResNeXt- & \multicolumn{3}{c}{ResNeXt-101-*}\\
      &   18 & 50 & 50-32x4 & 32x4 & 32x8 & 32x16\\ \midrule
Acc.  & 70.6 & 77.6 & 78.9 & 80.2 & 81.1 & 81.4\\
Gains & 0.0 & +1.2 & +1.3 & +1.7 & +2.0 & +1.8\\
\bottomrule
\end{tabular}}
\vspace{-5pt}
\caption{Self-training: top-1 accuracy of ResNet and ResNeXt models self-trained on the YFCC dataset. Gains refer to improvement over the fully supervised baseline.}
\label{tab:yfcc_selftrained_student_capacity}
\end{table}

\paragraph{Self-training: Ablation of the teacher/student.}
%
A special case of our proposed semi-supervised training approach is when the teacher and student models have the same architecture and capacity. In other words, the sampling is performed by the target model itself trained only on the available labeled dataset. In this case, our approach reduces to be a \emph{self-training} method.


%
Table~\ref{tab:yfcc_selftrained_student_capacity} shows the accuracy, with our default setting, 
for different models as well as the gain achieved over simply training on ImageNet for reference. All models benefit from self-training. Higher capacity models have relatively higher accuracy gains. For example, the ResNeXt-101 32x16 model is improved by 2\% compared to 1\% with ResNet-50. We also run a second round of self-training for ResNet-50 and ResNeXt-101 32x16 models and observed subsequent gains of 0.3\% and 0.7\%, respectively.

As discussed earlier, self-training is a challenging task since errors can get amplified during training.
Our experiments show that, with a fine-tuning stage to correct possible labeling noise, the model can improve itself by benefiting from a larger set of examples. However, by comparing these results to those provided in Table~\ref{tab:yfcc_student_capacity}, we observe that our teacher/student choice is more effective, as it generally offers a better accuracy for a specific target architecture.

\subsection{Parameter study}

\begin{figure*}
\begin{minipage}[t]{0.3\linewidth}
\centering
  \includegraphics[width=\linewidth]{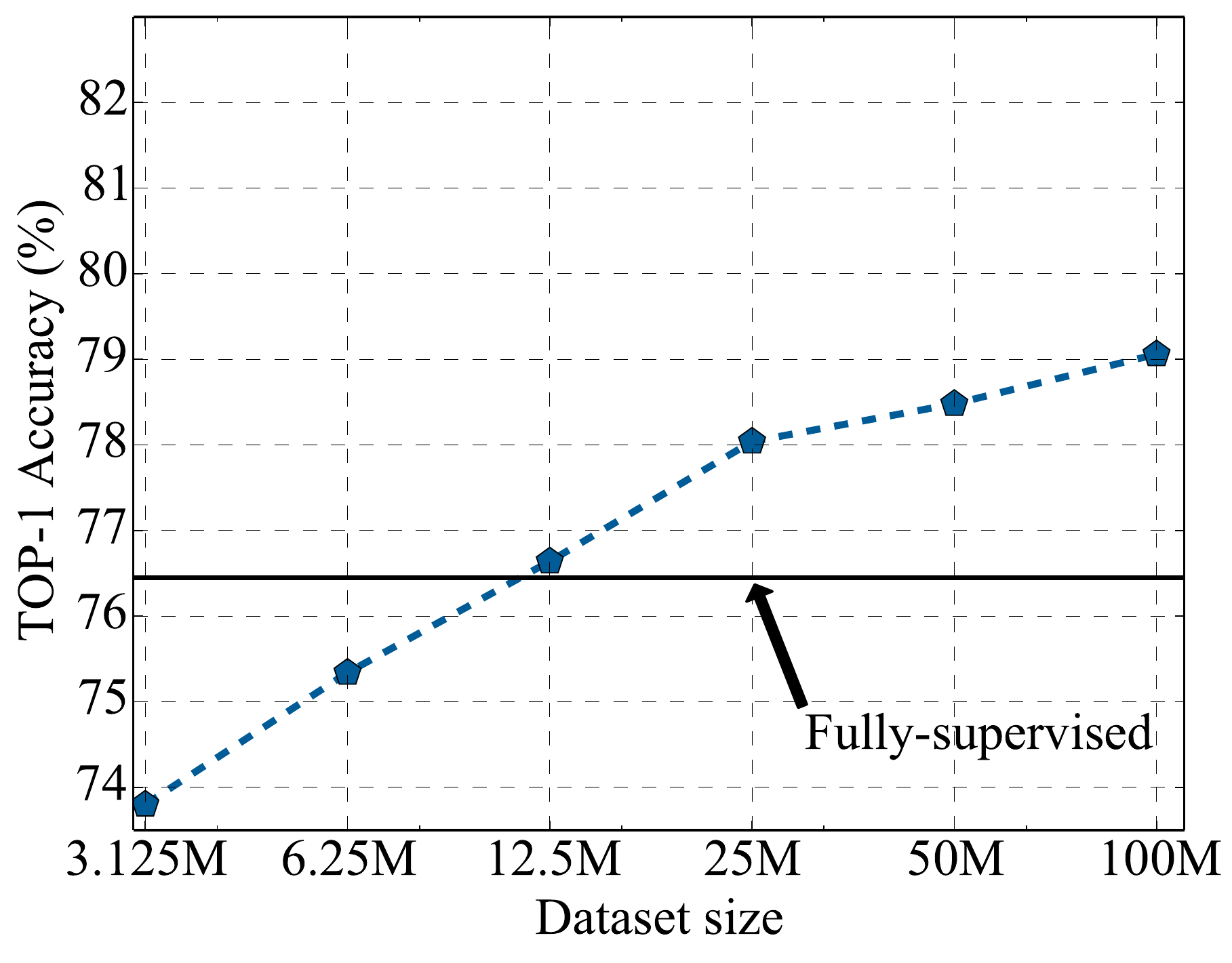}
  \caption{ResNet-50 student model accuracy as a function of the size of the unlabeled dataset $\mathcal U$. \label{fig:yfcc_db_size}}
\end{minipage}
\hfill
\begin{minipage}[t]{0.3\linewidth}
\centering
  \includegraphics[width=\linewidth]{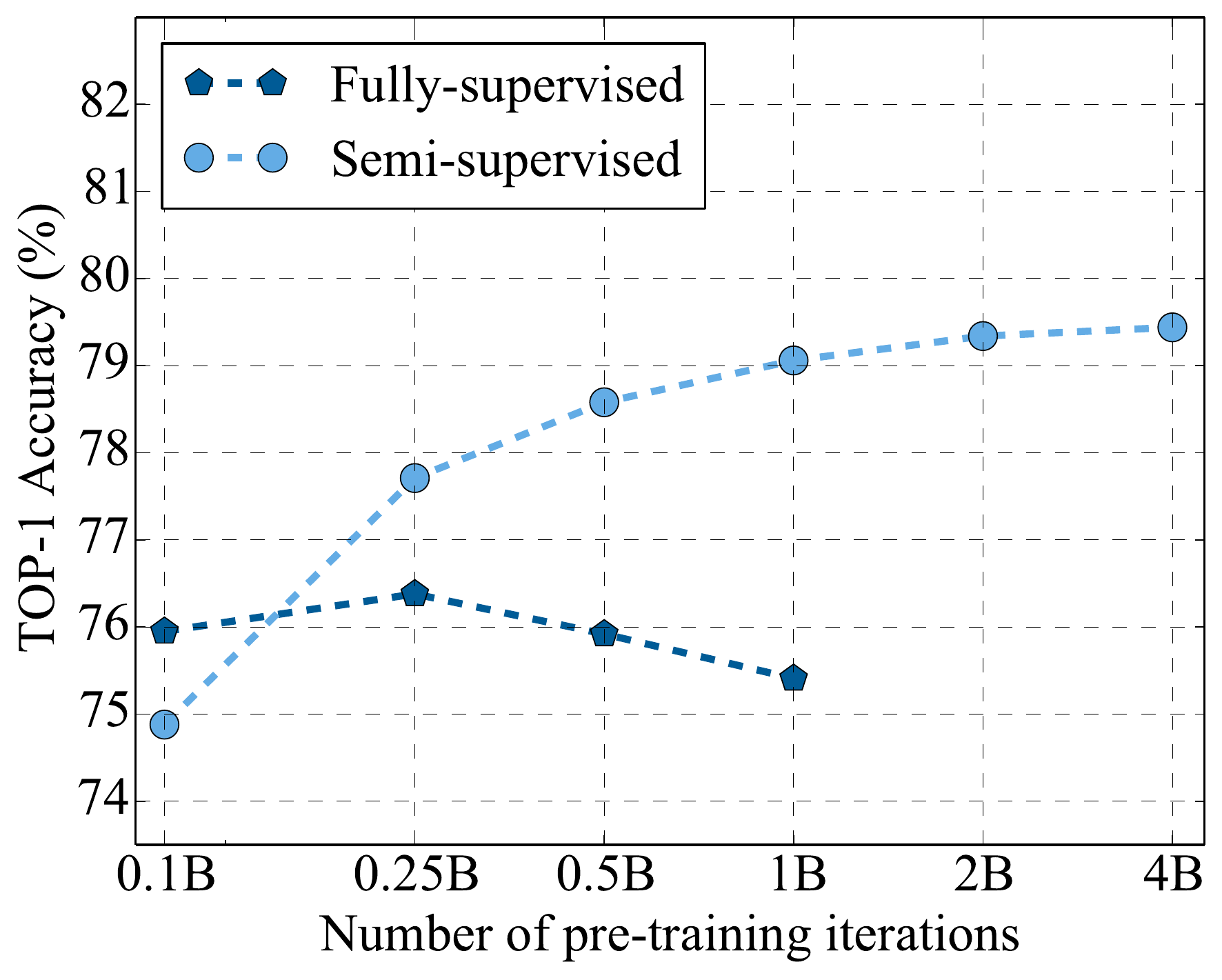}
  \caption{ Effect of number of training iterations on the accuracy of fully-supervised and semi-supervised ResNet-50 student models.
  \label{fig:effect_of_iters}}
\end{minipage}
\hfill
\begin{minipage}[t]{0.3\linewidth}
\centering
  \includegraphics[width=\linewidth]{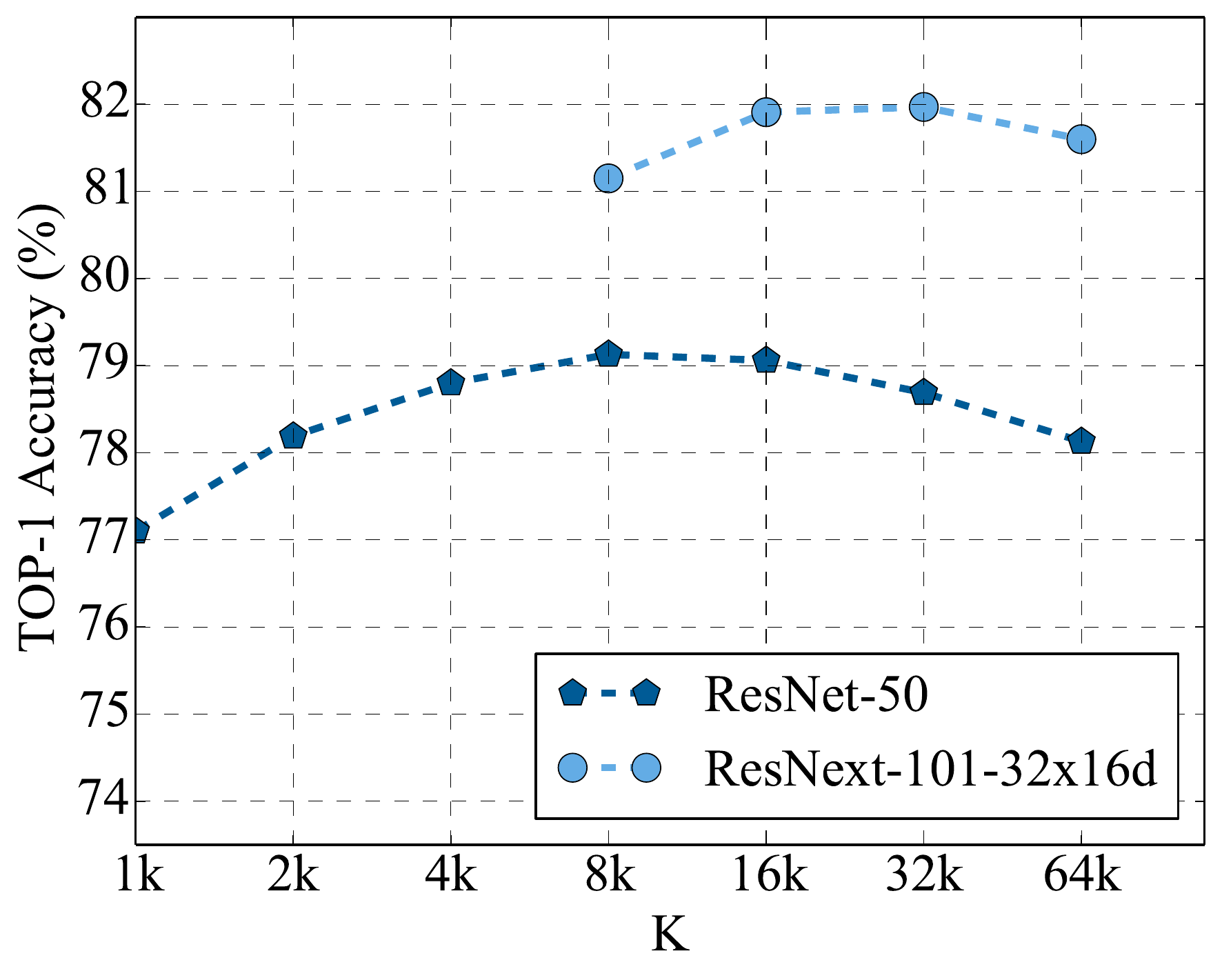}
  \caption{
  Student model accuracies as a function of the sampling hyper-parameter $K$.
\label{fig:yfcc_topk_ablations}}
\end{minipage}
\vspace{-10pt}
\end{figure*}

\paragraph{Size of the unlabeled set $\mathcal U$. }

A number of varying size datasets are created by randomly sampling from YFCC100M image distribution. The hyper-parameter value $K$ is adjusted for each random subset proportionally in order to marginalize the effect of the dataset scale. For example, values of $K=16k$, $8k$ and $4k$ are set for dataset sizes $100M$, $50M$ and $25M$, respectively. The semi-supervised model accuracy is plotted as a function of unlabeled dataset size in Figure~\ref{fig:yfcc_db_size}.
A fixed accuracy improvement is achieved every time the dataset size is doubled until reaching the dataset size of 25M.
However, this log-linear behavior disappears as the dataset size grows further. The target model is possibly reaching to its saturation point where adding additional training samples from the same YFCC image distribution does not help improve the accuracy.



Overall, the above experiments indicate that leveraging the large-scale of unlabeled data is important for good performance (Point 3 in our recommendations of Table~\ref{tab:recipe}).

\paragraph{Number of pre-training iterations. }Figure~\ref{fig:effect_of_iters} shows the performance as a function of total number of images processed during pre-training. To make sure that the learning rate drops by the same factor in each setting, we divide the overall training in to $13$ equally spaced steps and drop the learning rate by a factor of 2 after every step. We observe that performance keeps on improving as we increase the number of processed images. This indicates that our approach needs a longer training stage to be fully beneficial, as summarized by our Point~4 in Table~\ref{tab:recipe}). As a reference, we also show results of training a  ResNet-50 model on ImageNet with the same number of iterations. Increasing the number of training iterations did not help much and in fact reduced accuracy by a small margin. Note, we also tried the common choice of dropping the learning rate $3$ times by a factor of $0.1$, with similar conclusion.
Overall, we found that $1$ billion iterations offers a satisfactory trade-off between attaining a good accuracy and resource usage.

\paragraph{Parameters $K$ and $P$.} Figure~\ref{fig:yfcc_topk_ablations} shows the effect of varying the number $K$ of images selected per class for $P$\,=$\,10$, for ResNet-50 and ResNext-101-32x16 student models. For ResNet50, we observe that the performance first improves as we increase the value of $K$ to 8k due to increase in diversity as well as hardness of examples. It is stable in a broad 4k-32k regime indicating that a coarse sweep to find the optimal value is sufficient. Increasing $K$ further  introduces a lot of labeling noise in $\hat{\mathcal D}$ and the accuracy drops. Similar observations hold for ResNext-101-32x16 except that the higher capacity model can leverage more data. It achieves its best performance between 16k and 32k per class.

The performance does not vary much with $P$ for a near-optimal value of $K$ like 16k. A high value of $P$ produce a more balanced class distribution for pre-training by allowing a training example to appear in more than one ranked list, see the supplemental material for an analysis.
We fix $P=10$ in all of our experiments since it allows us to produces an almost balanced pre-training dataset (Point 5 in Table~\ref{tab:recipe}) when the collection size $\mathcal U$ is not large enough.

\subsection{Semi-weakly supervised experiments}
\label{sec:semiweakly}
Prior work~\cite{mahajan2018exploring} shows that images associated with meta-information semantically relevant to the task can improve performance. One of the ways to achieve this is to leverage task relevant search queries or hashtag
to create a noisy weakly-supervised dataset and use it as $\mathcal U$. Our ranking based selection could be viewed as a preliminary noise removal and class balancing process.

\begin{table}[t]
\centering {\small
\begin{tabular}{l|c|c|c}
\toprule
& Balanced & Selection & Accuracy \\ \midrule
balanced-ranked     & Yes & Ranked & 79.2 \\
unbalanced-ranked   & No & Ranked & 78.4\\
balanced-with-tags  & Yes & Random & 76.8\\ \midrule
supervised & - & - & 76.4\\
\bottomrule
\end{tabular}}
\vspace{-5pt}
\caption{Analysis of the selection step on IG-1B-Targeted. All methods selects a subset of 8 million images for training, our balanced-ranked method works the best. }
\label{tab:noise_bal_ablations}
\end{table}

\paragraph{Study: variants for $\hat{\mathcal D}$ selection procedure.} In order to investigate this further, we create three different $8$ million subsets of IG-1B-Targeted data to train a ResNet-50 student model: (1) {\it{balanced-ranked}} is created using our ranking approach with same number of examples (top-8k) per class, (2) {\it{unbalanced-ranked}} also uses our ranking approach but the number of examples per label follows a Zipfian distribution, and, (3) {\it{balanced-with-tags}} randomly selects $8k$ images per class using relevant hashtags. Please see supplementary material for details. Table~\ref{tab:noise_bal_ablations} shows the results of training on these datasets followed by fine-tuning on ImageNet. We observe that our classification-based selection via ranking is key to achieve a good performance. Note, leveraging a large amount of unlabeled data is necessary to obtain similar number of images per class.

\paragraph{Pre-training with hash tags.}
Pre-training on weakly-supervised data has recently achieved impressive performances. Motivated by these results, we follow Mahajan et. al. \etal~\cite{mahajan2018exploring} to train a ResNext-101 32x48 teacher model (85.4\% top-1 accuracy) by pre-training on IG-1B-Targeted and fine-tuning on ImageNet. Due to the larger size dataset, we use $K=64k$. Table~\ref{tab:weakly_sup_results} compares our method with a weakly-supervised approach~\cite{mahajan2018exploring} that pre-trains models on IG-1B-Targeted with hashtags as labels. This is a strong baseline with respect to the state of the art on most models. We observe a consistent improvement of 0.6\%--2.2\% with lower capacity models benefiting the most. All our models achieve the state-of-the-art performance with ResNet-50 achieving a solid $80.9\%$ accuracy. Hence, leveraging weakly-supervised data to pre-train the teacher model significantly improves the results (Point~6 in Table~\ref{tab:recipe}). 

\begin{table}[t]
\centering {\small
\begin{tabular}{@{}l|c|c@{\hspace{8pt}}c@{\hspace{8pt}}c@{}}
\toprule
& \multirow{2}{*}{ResNet-50} & \multicolumn{3}{@{}c@{}}{ResNeXt-101-*}\\
&     & 32x4 & 32x8 & 32x16\\
\midrule
Xie \etal~\cite{xie2017aggregated}            & 76.1 & 78.8 & - & -  \\
Mixup~\cite{zhang2017mixup}                   & 76.7 & 79.9 & - & - \\
LabelRefinery~\cite{bagherinezhad2018label}   & 76.5 & -    & - & - \\
Autoaugment~\cite{cubuk2018autoaugment}       & 77.6 & -    & - & - \\ \midrule
Weakly supervised~\cite{mahajan2018exploring} & 78.2 & 81.2 & 82.7 & 84.2\\
ours (semi-supervised)                        & 79.1 & 80.8 & 81.2 & 81.2 \\
ours (semi-weakly sup.)                 & 80.9 (\textbf{81.2}$^\dagger$) & \textbf{83.4} & \textbf{84.3} & \textbf{84.8}\\
\bottomrule
\end{tabular}}
\vspace{-5pt}
\caption{State of the art on ImageNet with standard architectures (ResNet, ResNext). The lower part of the table reports methods that leverage unlabeled data. Our teacher model is either learned on ImageNet (semi-supervised) or pre-trained on weakly-supervised IG-1B-Targeted (semi-weakly sup., $^\dagger$trained with \#iter=2B). The weakly supervised baseline~\cite{mahajan2018exploring} is learned on IG-1B-Targeted data followed by fine-tuning on ImageNet.}
\label{tab:weakly_sup_results}
\end{table}

\subsection{Comparison with the state of the art}

Table~\ref{tab:weakly_sup_results} compares our approach against the state of the art results for existing architectures, in particular the popular ResNet-50, for which we have several points of comparison.
As one can see, our approach provides a large improvement ($>$3\%) over all methods that only use labelled data. As discussed above, it also significantly outperforms (+0.6--2.5\%) a state-of-the-art weakly supervised method~\cite{mahajan2018exploring}.

\section{Other applications}
\label{sec:exp_video_transfer}
In this section, motivated by our findings on the ImageNet dataset, we apply our method to two different tasks: (1) video action classification, and, (2) transfer learning for improved classification using our semi-supervised image models.




\subsection{Video Classification Experiments} \label{ssec:video_exp}
We run experiments with the popular multi-class {\it{Kinetics}} video benchmark which has $\sim 246k$ training videos and $400$ human action labels. The models are evaluated on the $20k$ validation videos. Similar to IG-1B-Targeted, we construct an {\it{IG-Kinetics}} of $65$ million videos by leveraging $359$ hashtags that are semantically relevant to Kinetics label space. Please see supplementary material for details. We use R(2+1)D-d~\cite{Tran2017vid} clip-based models with model depths $18$ and $34$ and use RGB video frames.

The teacher is a weakly-supervised R(2+1)D-34 model with clip length $32$.
It is pre-trained with IG-Kinetics dataset and fine-tuned with labeled Kinetics videos.
We uniformly sample 10 clips from each video and average the softmax predictions to produce video level predictions. 
For our approach, we use $K$\,=\,4k and $P$\,=\,$4$ and IG-Kinetics as unlabeled data $\mathcal U$ to train student models. Please see the supplemental for training details.

Table~\ref{tab:vid} reports the results for $3$ different student models along with number of parameters and multiply-add FLOPS. For comparison, we also show results with: (1) training with full-supervision on Kinetics, and, (2) weakly-supervised pre-training on IG-Kinetics followed by fine-tuning on Kinetics. Our approach gives significant improvements over fully-supervised training. We also observe further gains over the competitive weakly-supervised pre-training approach with models having lower FLOPS benefiting the most. We also compare with other state-of-the-art approaches and perform competitively. We note that we process at most $32$ RGB frames of the input video (no optical flow), at a much lower resolution ($112\times112$) compared to the state of the art.

\begin{table}[t]
\centering {\small
\begin{tabular}{l|c|cc}
\toprule
Approach & Input & top-1 & top-5 \\ \midrule
\multicolumn{4}{c}{{{\it{R(2+1)D-18, clip length - 8, \# params 33M, FLOPS - 21B}}}}\\ \midrule
ours (semi-weakly sup.)    & RGB & 74.2 & 91.3 \\
fully-supervised    & RGB & 64.8 & 84.5 \\
weakly-supervised   & RGB & 71.5 & 89.7 \\ \midrule
\multicolumn{4}{c}{{{\it{R(2+1)D-18, clip length - 32, \# params 33M, FLOPS - 83B}}}}\\ \midrule
ours (semi-weakly sup.)       & RGB & 76.7 & 92.3 \\
fully-supervised    & RGB & 69.3 & 87.7 \\
weakly-supervised  & RGB & 76.0 & 92.0 \\ \midrule
\multicolumn{4}{c}{{{\it{R(2+1)D-34, clip length - 8, \# params 64M, FLOPS - 38B}}}}\\ \midrule
ours (semi-weakly sup.)       & RGB & 75.9 & 92.0 \\
fully-supervised    & RGB & 67.0 & 85.8 \\
weakly-supervised  & RGB & 74.8 & 91.2 \\ \midrule \midrule
NL I3D~\cite{Wang2018nonlocal}  & RGB & 77.7 & 93.3 \\
3 stream SATT~\cite{Bian2017kinetics} & RGB+flow+audio & 77.7 & 93.2 \\
I3D-Two-Stream~\cite{Carreira2017i3d} & RGB+flow & 75.7 & 92.0 \\
\bottomrule
\end{tabular}}
\vspace{-7pt}
\caption{Accuracy on Kinetics video dataset for different approaches using R(2+1)D models. {\it{Weakly-supervised}} refers to weakly-supervised training on IG-Kinetics followed by fine-tuning on Kinetics. Our approach uses R(2+1)D-34 teacher model with clip length $32$.
\label{tab:vid}}
\vspace{-5pt}
\end{table}


\subsection{Transfer Learning Experiments} \label{ssec:representation_learning}

\begin{table}[t]
\centering {\small
\begin{tabular}{@{}l|c|c|c|c@{}}
\toprule
Pre-trained    & ImageNet   & weakly  &  semi        & semi-weakly  \\
Model          & sup.  & sup. & sup. (ours)  & sup. (ours)   \\ \midrule
{\it{full-ft}} &  82.1 & 83.2 & 83.6 &  {\bf{84.8}} \\
{\it{fc-only}} &  73.3 & 74.0 & 80.4 &  {\bf{80.7}} \\
\bottomrule
\end{tabular}}
\vspace{-7pt}
\caption{CUB2011: Transfer learning accuracy (ResNet50).
\label{tab:cub_finetune_exp}}
\vspace{-20pt}
\end{table}

ImageNet has been used as the de facto pre-training dataset for transfer learning.
A natural question that arises is how effective models trained using our approach are for transfer learning tasks.
We consider CUB2011~\cite{Wah2011CUB} dataset with 5994 training images and 5794 test images associated with 200 bird species (labels).
Two transfer learning settings are investigated: (1) {\it{full-ft }} involves fine-tuning the full network, and, (2) {\it{fc-only }} involves extracting features from the final fc layer and training a logistic regressor.

Table~\ref{tab:cub_finetune_exp} reports the accuracy of ResNet-50 models trained with our semi-supervised (Section~\ref{sec:stepanalysis}) and semi-weakly supervised (Section~\ref{sec:semiweakly}) methods to the performance of fully-supervised and weakly-supervised models. See the supplemental for the training settings. The models trained with our approach perform significantly better. Results are particularly impressive for {\it{fc-only}} setting, where our semi-weakly supervised model outperforms highly competitive weakly-supervised model by $6.7\%$.

\section{Conclusion}
\label{sec:conclusion}
We have leveraged very large scale unlabeled image collections via semi-supervised learning to improve the quality of vanilla CNN models. The scale of the unlabelled dataset allows us to infer a training set much larger than the original one, allowing us to learn stronger convolutional neural networks. Unlabelled and labelled images are exploited in a simple yet practical and effective way, in separate stages.
An extensive study of parameters and variants leads us to formulate recommendations for large-scale semi-supervised deep learning. As a byproduct, our ablation study shows that a model self-trained with our method also achieves a compelling performance.
Overall, we report state-of-the-art results for several architectures.


{\small
\bibliographystyle{ieee}
\bibliography{self_training}}

\clearpage



\definecolor{lightgrey}{rgb}{0.97, 0.97, 0.97}
\definecolor{darkgreen}{RGB}{0, 140, 0}
\definecolor{antiquefuchsia}{rgb}{0.57, 0.36, 0.51}
\definecolor{auburn}{rgb}{0.43, 0.21, 0.1}

\setlist[itemize]{%
labelsep=5pt,%
labelindent=0.4\parindent,%
itemindent=0pt,%
leftmargin=*,%
itemsep=4pt,
topsep=0pt,
parsep=0pt,
partopsep=0pt
}

\setlist[enumerate]{%
labelsep=5pt,%
labelindent=0.4\parindent,%
itemindent=0pt,%
leftmargin=*,%
itemsep=3pt,
topsep=4pt,
parsep=0pt,
partopsep=0pt
}


\makeatletter
\renewcommand{\paragraph}{%
  \@startsection{paragraph}{4}%
  {\z@}{0.6em}{-1em}%
  {\normalfont\normalsize\bfseries}%
}


\iccvfinalcopy 

\def\iccvPaperID{5006} 
\def\httilde{\mbox{\tt\raisebox{-.5ex}{\symbol{126}}}}

\ificcvfinal\pagestyle{empty}\fi







\section*{SUPPLEMENTAL MATERIAL}
\addcontentsline{toc}{section}{SUPPLEMENTAL MATERIAL}

This supplemental material provides details that complement our paper. First we provide statistics on our parameter $P$ in Section~\ref{sec:parameterP}. Then Section~\ref{sec:semiweakly} provides details about the variants discussed in Section 4.4. Section~\ref{sec:selftrained} provides additional results that show the effectiveness of our method to train a teacher with our own technique. Section~\ref{sec:videos} describes how we collected our IG-Kinectics video dataset. Section~\ref{sec:dedup} explains our de-duplication procedure that ensures that no validation or test data belongs to our unlabelled dataset. Finally, we give in Section~\ref{sec:transfercub} the parameters associated with our transfer learning experiments on CUB2011.

\setcounter{section}{0}
\section{Effect of parameter $P$}
\label{sec:parameterP}
Table \ref{tab:yfcc_p_ablations} shows the effect of varying $P$ on the ResNet-50 student model trained with the vanilla ResNext-101 32x48 teacher model for $K=16k$. We consider YFCC-100M and construct two datasets of $25M$ and $100M$ examples by randomly sampling the data. We notice that increasing $P$ balances the ranked list sizes across all the classes (column $| \hat{\mathcal D} |$). We also observe that it does not have a significant impact on the performance. 
Since our main focus of our work is large-scale training with an interest to handle under-represented classes, we prefer to guarantee a fixed number of samples per class and therefore choose $P=10$ in all the experiments.

\begin{table}[h]
\centering {\small
\begin{tabular}{c|c|c|cc}
\toprule
P  & $| \mathcal{U} |$ & $| \hat{\mathcal D} |$  & top-1 \\ \midrule 
1  & 25M   & 10.1M    & 78.1 \\
3  & 25M   & 14.3M    & 78.1  \\
5  & 25M   & 15.5M    & 77.9 \\
10 & 25M   & 16.0M    & 77.9  \\  \midrule 
1  & 100M  & 14.4M    & 79.1  \\  
3  & 100M  & 15.9M    & 78.9  \\  
5  & 100M  & 16.0M    & 79.2  \\  
10 & 100M  & 16.0M    & 79.1  \\  
\bottomrule
\end{tabular}}
\caption{The effect of the training hyper-parameter P on the accuracy of the ResNet-50 student model.}
\label{tab:yfcc_p_ablations}
\end{table}

\section{Semi+weakly supervised: variants for $\hat{\mathcal D}$}
\label{sec:semiweakly}
This section gives more details about the variants for collecting a training set, as discussed in Section 4.4. 
We create three different 8 million subsets of IG-1B-Targeted data.
\begin{itemize}
\item {\it{balanced-ranked}} is created using our ranking approach with same number of examples (top-8k) per class.
\item {\it{unbalanced-ranked}} also uses our ranking approach but the number of examples per label follows a Zipfian distribution. In order to make sure that our Zipfian distribution matches the real-world data, we assign the $1.5k$ hashtags from IG-1B-Targeted to one of ImageNet classes to determine the approximate number of images per class. We then use this distribution as basis to determine $K$ to select $top-K$ images for each class so that total number of images across the classes is $8M$.
\item {\it{balanced-with-tags}} randomly selects $8k$ images per class using relevant hashtags after mapping them to ImageNet classes and does not use any ranked lists.
\end{itemize}

\begin{table}[t]
\centering {\small
\begin{tabular}{l|c|c} 
\toprule
Student & \multirow{2}{*}{top-1}  & \multirow{2}{*}{Gain (\%)} \\
Model  &    &  \\
\midrule
ResNet-18            & 72.8  & +0.2 \\
ResNet-50            & 79.3  & +0.2 \\
ResNext-50-32x4      & 80.3  & +0.4 \\
ResNext-101-32x4     & 81.0  & +0.2 \\
ResNext-101-32x8     & 81.7  & +0.5 \\
ResNext-101-32x16    & 81.9  & +0.7 \\
\bottomrule
\end{tabular}}
\vspace{-5pt}
\caption{The top-1 accuracy gain (\%) of varying student models pre-trained with a ``self-trained'' ResNext-101-32x16d teacher model on YFCC dataset. The gain is the absolute accuracy improvement over the semi-supervised result in Table 2 in the main paper.}
\label{data:yfcc_selfco_trained_student_capacity}
\end{table}

\section{Self-trained teacher models}
\label{sec:selftrained}

In the paper, we observe that in the self-training setting where teacher and student models have the same architecture and capacity, we still get $1\%-2\%$ gains in accuracy over the fully-supervised setting. Motivated by this, we consider another scenario where we take a high capacity ResNext-101-32x16 model trained on ImageNet and self-train it using YFCC-100M as unlabeled data. \textbf{We then use this self-trained model as a teacher} and train a lower capacity student model using our approach on YFCC-100M.
Standard training settings mentioned in the main paper are used. Table~\ref{data:yfcc_selfco_trained_student_capacity} shows the accuracy of different student models trained in the above setting. For reference, we also compare against semi-supervised result from Table 2 in the main paper using an ImageNet trained ResNext-101 32x48 teacher model. We observe a consistent $0.2\%-0.7\%$ improvement even by using a lower capacity self-trained ResNeXt-101 32x16 teacher. {\bf{We intend to release these models publicly.}}

\begin{table}[t]
\centering {\small
\begin{tabular}{c|c|l|c} 
\toprule
\multirow{2}{*}{Model} & Clip & Minibatch & Initial\\
& len. & size & LR\\
\midrule
R(2+1)D-18 & 8 & $128 \times 16 = 2048$   &  $0.064/256 \times 2048$ \\
R(2+1)D-18 & 32 & $128 \times 8 = 1024$  &  $0.064/256 \times 1024$ \\
R(2+1)D-34 & 8  & $128 \times 16 = 2048$  &  $0.064/256 \times 2048$ \\
R(2+1)D-34 & 32  & $128 \times 8 = 1024$ &   $0.064/256 \times 1024$ \\
\bottomrule
\end{tabular}}
\vspace{-5pt}
\caption{Hyper-parameters for pre-training video models.}
\label{tab:vid_prestats}
\end{table}

\section{Weakly-supervised video experiments}
\label{sec:videos}
\begin{table}[t]
\centering {\small
\begin{tabular}{c|c|c|c|c} 
\toprule
Depth, & Initial & Total & warm & LR\\
Clip len. & LR & epochs & up & steps\\
\midrule
\multicolumn{5}{c}{fully-supervised}\\ \midrule
18,8  &  $1.6\mathrm{e}{-1}/256 \times 2048$ & 180 & 40 & $[40]\times 3$ \\
18,32 & $3.2\mathrm{e}{-1}/256\times 1024$  & 180 & 40 & $[40]\times 3$ \\
34,8  & $1.6\mathrm{e}{-1}/256 \times 2048$  & 180 & 40 & $[40]\times 3$ \\ \midrule
\multicolumn{5}{c}{weakly-supervised}\\ \midrule
18,8  &  $4.8\mathrm{e}{-4}/256 \times 2048$ & 100 & 16 & $[28]\times 3$ \\
18,32 & $9.6\mathrm{e}{-4}/256\times 1024$  & 72 & 16 & $[16]\times 3$ \\
34,8  & $4.8\mathrm{e}{-4}/256 \times 2048$  & 72 & 16 & $[16]\times 3$ \\ \midrule
\multicolumn{5}{c}{semi-weakly supervised}\\ \midrule
18,8  &  $4.8\mathrm{e}{-4}/256 \times 2048$ & 72 & 16 & $[16]\times 3$ \\
18,32 & $9.6\mathrm{e}{-4}/256\times 1024$  & 72 & 16 & $[16]\times 3$ \\
34,8  & $4.8\mathrm{e}{-4}/256 \times 2048$  & 72 & 16 & $[16]\times 3$ \\
\bottomrule
\end{tabular}}
\vspace{-5pt}
\caption{Hyper-parameters for fully-supervised training and fine-tuning video models. }
\label{tab:vid_finestats}
\end{table}
\subsection{Construction of IG-Kinetics dataset}
We consider millions of public videos from a social media and the associated hashtags. For each label in Kinetics dataset, we take the original and stemmed version of every word in the label and concatenate them in different permutations. For example, for the label {\it{"playing the guitar"}}, different permutations are - {\it{"playingtheguitar"}}, {\it{"playguitar"}}, {\it{"playingguitar"}}, {\it{"guitarplaying"}}, etc. We then find all the hashtags and associated videos that match any of these permutations and assign them the label. Finally, we end up with $65$ million videos with $369$ labels. There are no corresponding hashtags for $31$ labels.
\subsection{Training details}
We first down-sample the video frames to a resolution of $128 \times 171$ and generate each video clip by cropping a random $112 \times 112$ patch from a frame. We also apply temporal jittering to the input. We train our models using synchronous stochastic gradient descent (SGD) on $128$ GPUs across $16$ machines. For the video clip with 32 frames, each GPU processes $8$ images (due to memory constraints) at a time and apply batch normalization~\cite{ioffe2015batch} to all convolutional layers on each GPU. For the 8 frames video clip, each GPU processes $8$ images at a time. The weight decay parameter is set to $0.0001$ in all the experiments. We set the learning rate following the linear scaling procedure proposed in~\cite{goyal2017imagenethour} with a warm-up.
\paragraph{Pre-training: } Both weakly-supervised and semi-weakly supervised settings require pre-training the models on IG-Kinetics. We process 490M videos in total. Table~\ref{tab:vid_prestats} shows the minibatch size and learning rate settings for different models. We decrease the learning rate by a factor of $2$ at equally spaced steps such that the total number of reductions is $13$ over the course of training.
\paragraph{Fine-tuning and fully-supervised setting: }Table~\ref{tab:vid_finestats} shows the training settings for fine-tuning and fully-supervised runs. For fine-tuning experiments, we do a grid-search on the initial learning rate and LR schedule using a separate held-out set. Learning rate decay is set to $0.1$ for all the experiments.

\section{Deduplication of images}
\label{sec:dedup}
In order to ensure a fair evaluation in our large-scale experiments, it is crucial that we remove images in the large-scale dataset that are also present in the labeled test or validation set. We leverage ImageNet trained ResNet-18 model and use its $512$ dimensional {\it{pool5}} features to compute the Euclidean distance between the images from ImageNet validation set and YFCC-100M after L2 normalization. We use the Faiss~\cite{FAISS} nearest neighbor library to implement the searchable image index of 100 million images. We sorted all pairs of images globally according to their Euclidean distances in ascending order and manually reviewed the top pairs. The top ranking $5000$ YFCC images are removed from the dataset. The same procedure is applied to CUB2011 to remove $218$ YFCC-100M images.

\begin{table}[t]
\centering {\small
\begin{tabular}{c|c|c|c} 
\toprule
Initial & Total & LR  & Weight\\
 LR & Epochs & Steps & decay\\
\midrule
$0.0025/256 \times 768$ & 300 & $[100] \times 3$ & 0.001 \\
\bottomrule
\end{tabular}}
\vspace{-5pt}
\caption{Hyper-parameters for full fine-tuning of models on CUB2011 dataset. Same settings are used for different kinds of supervision (Table 8 in main paper) and models.}
\label{tab:cub_finestats}
\end{table}
\section{Transfer learning on CUB2011: parameters}
\label{sec:transfercub}
For the full fine-tuning scenario, we fine-tune the models on $32$ GPUs across $4$ machines. Table~\ref{tab:cub_finestats} provides the parameter settings for different runs.

\end{document}